\documentclass[runningheads]{llncs}
\usepackage{graphicx}
\usepackage{graphicx}
\usepackage[gen]{eurosym}
\usepackage{amsmath, amssymb, bm}
\usepackage{url}
\usepackage{subfigure}
\usepackage{bold-extra}
\usepackage[normalem]{ulem}
\usepackage{wrapfig}
\usepackage{array,multirow}
\usepackage{tikz}
\usepackage[misc]{ifsym}
\usepackage{hyperref}
\newcommand*\circled[1]{\tikz[baseline=(char.base)]{
  \node[shape=circle,draw,inner sep=1pt] (char) {#1};}}

\begin{document}
\title{Beyond Gut Feel: Using Time Series Transformers to Find Investment Gems}
\titlerunning{Beyond Gut Feel: Using Time Series Transformers to Find Investment Gems}
%
\author{
Lele Cao$^{*\,\textrm{\Letter}}$\,\inst{1}\orcidID{0000-0002-5680-9031} \and
Gustaf Halvardsson\thanks{Gustaf Halvardsson (currently works for BCG GAMMA, part of BCG X) contributed (equally as Lele Cao) to this work as part of his Master's Thesis project carried out in EQT Motherbrain supervised by Lele Cao (industrial supervisor) and Pawel Herman (academic supervisor). \;\;\; \textsuperscript{\textrm{\Letter}} Corresponding author: \href{mailto:caolele@gmail.com}{Lele Cao}.}\inst{1,2}\orcidID{0000-0002-6559-1521} \and
Andrew McCornack\inst{1}\orcidID{0009-0000-4516-0137} \and
Vilhelm von Ehrenheim\inst{1,3}\orcidID{0000-0002-4210-4989} \and
Pawel Herman\inst{2}\orcidID{0000-0001-6553-823X}
}
\authorrunning{L. Cao and G. Halvardsson et al.}
%
\institute{
Motherbrain, EQT Group, Regeringsgatan 25, 11153 Stockholm, Sweden \\
\email{caolele@gmail.com}\;\;\;\;\;
\email{gustaf.halvardsson@gmail.com}\;\;\;\;\;
\email{andrew.mccornack@eqtpartners.com}\;\;\;\;\;
\email{vonehrenheim@gmail.com}\and
KTH Royal Institute of Technology, Lindstedtsvagen 5, 11428 Stockholm, Sweden
\email{paherman@kth.se}\and
QA.tech, Gävlegatan 16, 11330 Stockholm, Sweden
}
\maketitle              
\begin{abstract}
This paper addresses the growing application of data-driven approaches within the Private Equity (PE) industry, particularly in sourcing investment targets (i.e.,~companies) for Venture Capital (VC) and Growth Capital (GC). We present a comprehensive review of the relevant approaches and propose a novel approach leveraging a Transformer-based Multivariate Time Series Classifier (TMTSC) for predicting the success likelihood of any candidate company. The objective of our research is to optimize sourcing performance for VC and GC investments by formally defining the sourcing problem as a multivariate time series classification task. We consecutively introduce the key components of our implementation which collectively contribute to the successful application of TMTSC in VC/GC sourcing: input features, model architecture, optimization target, and investor-centric data processing. Our extensive experiments on two real-world investment tasks, benchmarked towards three popular baselines, demonstrate the effectiveness of our approach in improving decision making within the VC and GC industry.

\keywords{Company success prediction \and Venture capital \and Growth equity \and Private equity \and Investment \and Multivariate time series.}
\end{abstract}
\section{Introduction}
Private Equity (PE) is a rapidly growing segment of the investment industry that manages funds on behalf of institutional and accredited investors. PE firms acquire and manage companies with the goal of achieving high, risk-adjusted returns through subsequent sales \cite{cao-etal-2023-ascalable}. These acquisitions can involve majority shares of private or public companies, or investments in buyouts as part of a consortium.
Common PE investment strategies, as identified by \cite{block2019private}, include Venture Capital (VC), Growth Capital (GC), and Leveraged Buyouts (LBO). These strategies offer varying degrees of risk and return potential, depending on the investment objectives and time horizon of the PE fund.
The ability to accurately assess the likelihood of company success is crucial for PE firms in identifying attractive investment targets. 
Traditional evaluation of company performance often relies on manual analysis of financial statements or proprietary information, which may not be sufficient for capturing the dynamic nature of companies, especially those in early-stage or high-growth industries.  This evaluation approach is often time consuming and as a result not every potential company can be properly evaluated.
Therefore, there is a growing interest in leveraging data-driven methods to (1) debias decisions, so that the individual investment decision made for a particular deal is expected to drive lower risk and higher ROI (return on investment); and (2) enable automation, so that more companies can be evaluated without the need for additional resources \cite{cao2022using}.

For LBO in the PE industry, data-driven approaches may be less relevant due to the combination of two reasons:
(1) LBO professionals often track and maintain in-depth knowledge of late-stage companies\footnote{Generally, a company is considered late-stage when it has proven that its concept and business model work, and it is out-earning its competitors.} in a few focus sectors, resulting in unique knowledge and understanding that can hardly be entirely replaced by public (or even proprietary) data; 
(2) the number of LBO investments is usually less than VC and GC leading to a lower sourcing frequency. 
VC investments often involve early-stage companies with prone-to-change business models and limited revenue, making data-driven approaches valuable for evaluating their growth potential. 
Additionally, VC investors typically manage larger portfolios with higher investment frequency, necessitating the use of data-driven models for efficient decision-making in identifying and evaluating investment opportunities.
In practice, historical financial data (e.g.,~revenue) of startup\footnote{A startup is a dynamic, flexible, high risk, and recently established company that typically represents a reproducible and scalable business model. It provides innovative products or services, and has limited funds and resources \cite{j8_santisteban2021critical,j22_skawinska2020success,blank2013lean,cao2022using}.} or scaleup\footnote{A startup moves into scaleup territory after proving the scalability and viability of its business model and experiencing an accelerated cycle of revenue growth. 
This transition is usually accompanied by the fundraising of outside capital \cite{cavallo2019fostering}.} companies are commonly perceived as a good approximation of their true valuations \cite{cao-etal-2022-sire}.
The financial information of GC targets (scaleups) is much more accessible than that of VC targets (startups). 
Therefore, GC practitioners often directly use financial metrics to calculate the company's valuation for sourcing, which is why the adoption of big data in GC sourcing may not be as intensive as in VCs.
However, data-driven approaches may still provide additional insights in assessing the growth potential and financial performance of the GC targets.

Our contributions significantly advance data-driven strategies for sourcing investment opportunities in the VC and GC sectors by predicting the potential success of companies. These advancements include:
\begin{itemize}
\item We formally define the sourcing problem for VC/GC investments as a multivariate time series classification task and propose to employ a Transformer-based Multivariate Time Series Classifier (TMTSC) to address it.
\item We introduce key components of our implementation, including input features, model architecture and optimization target, which all contribute to the successful application of TMTSC in VC/GC sourcing.
\item We carry out extensive experiments, comparing TMTSC with widely adopted baselines on two real-world tasks, and demonstrate the effectiveness of our approach using a diverse set of evaluation metrics and strategies.
\end{itemize}

\section{Related Work}\label{sec:related-work}
Over the past two decades, data-driven approaches have been dominating research on deal sourcing for VC, i.e.~identifying startups that eventually turn into unicorns\footnote{
Unicorn and near-unicorn startups are private, venture-backed firms with a valuation of at least \$500 million at some point \cite{chernenko2021mutual}.
}. 
In recent years, however, research has begun to intensify on GC deal sourcing, transforming the way scaleup companies are identified and assessed.
Based on our extensive literature survey, data-driven methods for VC/GC deal sourcing can be broadly categorized into Statistical and Analytical (S\&A) methods, conventional Machine Learning (ML) methods, and Deep Learning (DL) methods. 
S\&A work
\cite{j33_malmstrom2020they,gompers2020venture,j30_kaiser2020value,j8_santisteban2021critical} typically starts with defining some hypotheses followed by testing them using statistical tools.
However, developing effective hypotheses for S\&A approaches is a challenging task that requires simplicity, conciseness, precision, testability, and most importantly, a grounding in existing literature or established theory, as emphasized in \cite{williamson2002research}.
It is worth mentioning that while DL methods technically fall under the broader umbrella of ML, we discuss DL work separately in recognition of its increasing popularity and relevance to our research.

\vspace{-6pt}
\subsection{Conventional Machine Learning Methods}
\label{sec:rw-cml-methods}
Over the last few years, there has been a growing interest in leveraging ML algorithms for {\it hypothesis mining} from data, as an alternative to manually defining hypotheses upfront. 
Hypothesis mining involves conducting explainability analysis on trained ML models to summarize, rather than explicitly define, hypotheses \cite{w5_guerzoni2019survival}. 
For instance, by training an ML model on a labeled dataset containing features of various companies, and quantifying how changes in these features impact the prediction target (i.e.~success probability), one can distill hypotheses that describe the relationships between the relevant features and the prediction target. 
Compared to S\&A, hypothesis mining is a much more structured procedure that trains an ML model using the entire dataset at hand.
In general, ML based approaches, as demonstrated in previous works such as \cite{krishna2016predicting,arroyo2019assessment,j30_kaiser2020value,j17_bonaventura2020predicting,zbikowski2021machine}, typically require practitioners to define the input data $\mathbf{x}$ and annotation $y$ (labeling ``good'' or ``bad'' investment according to some criteria) before training a model $f(\cdot)$ that maps $\mathbf{x}$ to $y$, i.e.,~$y=f(\mathbf{x})$.
With the rapid growth of dataset size and diversity (origin and modality), conventional ML models\footnote{The frequently applied conventional ML models include many such as decision tree~\cite{arroyo2019assessment}, random forest~\cite{krishna2016predicting}, logistic regression~\cite{j30_kaiser2020value}, and gradient boosting~\cite{zbikowski2021machine}.} sometimes struggle to fit the large and unstructured\footnote{Unstructured data, such as image and timeseries, is a collection of many varied types that maintains their native form, while
structured data is aggregated from original (raw) data and is usually stored in a tabular form.} data due to lack of {\it capacity} and {\it expressivity}\footnote{\label{footnote:capacity-expressivity}{\it Expressivity} describes the classes of functions a model can approximate, and {\it capacity} measures how much ``brute force'' the model has to fit the data.}.

\vspace{-6pt}
\subsection{Deep Learning Methods}\label{sec:rw-dl-methods}
Most recently, DL algorithms have attracted an increasing number of researchers hunting for good VC/GC investment targets. 
DL is implemented (entirely or partly) with ANNs (artificial neural networks) that utilize at least two hidden layers of neurons.
The {\it capacity} of DL can be controlled by the number of neurons (width) and layers (depth) \cite{goodfellow2016deep}.
Deep ANNs are exponentially {\it expressive} with respect to their depth \cite{raghu2017expressive}.
While structured data is commonly used in many DL methods, such as \cite{c9_dellermann2021finding,j26_bai2021startup,b1_ang2022using}, unstructured data is increasingly recognized as an important complement to structured data in recent studies \cite{t1_stahl2021leveraging,t2_horn2021deep,w1_lyu2021graph,w3_gastaud2019varying,c16_chen2021trend}, or even as a standalone input to the model \cite{c1_zhang2021scalable,j9_tang2022deep}.
Unstructured data often contains large-scale and intact-yet-noisy signals, which may result in superior performance when a proper DL approach is applied \cite{w6_garkavenko2022you}.

The main types of unstructured data seen include text~\cite{c16_chen2021trend}, graph~\cite{j16_allu2022predicting}, image~\cite{c21_cheng2019success}, video~\cite{j9_tang2022deep}, audio~\cite{j43_shi2021leveraging} and time series~\cite{c16_chen2021trend}.
Among these, fine-grained multivariate time series, which encompass various aspects of a company over time, hold particular significance for deal sourcing in the VC/GC domain. Some examples of these aspects include financial performance, team dynamics, funding rounds, market conditions, and other key indicators.
Especially for GC, financial time series become highly relevant for evaluating scaleup companies whose periodical financial data points are usually available to the potential investor \cite{cao-etal-2022-sire}. 
Due to the proprietary, costly, and scarce nature of multivariate time series company data, there is a limited number of DL based approaches in the literature that utilize time series as model input. 
To the best of our effort, we identified only three such studies \cite{c16_chen2021trend,t1_stahl2021leveraging,t2_horn2021deep}, highlighting the challenges associated with utilizing multivariate time series data to source investment targets for Venture and Growth Capital.
Inspired by \cite{zerveas2021transformer}, we frame the problem as a multivariate time series classification task and propose a solution that leverages a Transformer model. 
Our approach also incorporates carefully designed input features, optimization target, and investor-centric data processing \cite{cao2022using}.

\begin{figure*}[t]
\centering
  \includegraphics[width=\textwidth]{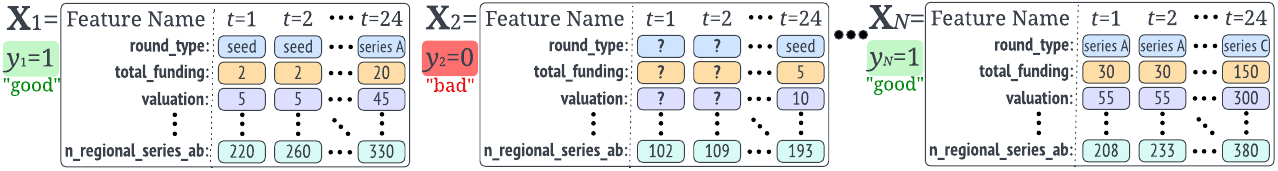}
  \vspace{-10pt}
  \caption{An illustration of multivariate time series dataset ($N$ samples) to train models for VC and GC sourcing.}
  \label{fig:teaser}
  \vspace{-10pt}
\end{figure*}

\section{The Approach}
\label{sec:approach}
Our approach tackles the problem of identifying good investment targets for VC and GC by framing it as a multivariate time series classification task. 
Specifically, each potential investment target (i.e.~candidate company) is represented by a multivariate time series $\mathbf{X}\in\mathbb{R}^{T\times K}$, as shown in Figure~\ref{fig:teaser}. 
$\mathbf{X}$ consists of $T$ observations, each containing $K$ variables that describe different aspects (e.g.,~funding, revenue, etc.) of the corresponding company. Formally, each sample $\mathbf{X}$ is a sequence of $T$ feature vectors: $\mathbf{X}\!=\![\mathbf{x}_1, \mathbf{x}_2, \ldots, \mathbf{x}_t, \ldots, \mathbf{x}_T]$, where $\mathbf{x}_t\!\in\!\mathbb{R}^K$. At each time step $t$, we collect $K$ numerical or categorical features about the company to form the vector $\mathbf{x}_t$, which captures a multi-view snapshot of the company at that time point. The last vector $\mathbf{x}_T$ represents the most recent state of the company. 
Depending on the data available, the time interval between two adjacent time points, $t$ and $t+1$, can be set to a month, a quarter, or any other length of choice. 
By adopting this representation, we can model a multi-view evolution of each company over time and make informed predictions about their future success.

We collect a set of $N$ samples, each corresponding to a company, denoted by ${\mathbf{X}_1, \mathbf{X}_2, \ldots, \mathbf{X}_n, \ldots, \mathbf{X}_N}$.
For each sample $\mathbf{X}_n$, we have a binary ground truth label $y_n\!\in\!\{0,1\}$ indicating a ``bad'' or ``good'' investment target according to some criteria.
Details of how we define and collect these labels are explained in Section~\ref{sec:opt-target}.
We construct a dataset $\mathfrak{U}$ from these samples and labels as $\mathfrak{U}=\{(\mathbf{X}_1, y_1), (\mathbf{X}_2, y_2), \ldots, (\mathbf{X}_N, y_N)\}$, where $n\in\mathbb{Z}\cap[1,N]$.
Our objective then is to {\bf train a model on $\mathfrak{U}$ to accurately predict the ground truth labels $y_n$ using $\mathbf{X}_n$}.
We use $\hat{y}_n\in[0,1]$ to denote the predicted probability of future success of the company represented by $\mathbf{X}_n$, in order to distinguish it from the ground truth label $y_n$.
For the sake of brevity, we use general terms $\mathbf{X}$, $y$, and $\hat{y}$ to denote $\mathbf{X}_n$, $y_n$, and $\hat{y}_n$, respectively.

\subsection{Time Series Features}
\label{sec:ts-features}
We define the input time series features $\mathbf{X}$ by constructing 16 time series that fall into 6 feature categories, as summarized in \cite{cao2022using}. These categories are \circled{\small\bf1} {\bf funding}, \circled{\small\bf2} {\bf founder/owner}, \circled{\small\bf3} {\bf team}, \circled{\small\bf4} {\bf investor}, \circled{\small\bf5} {\bf web}, and \circled{\small\bf6} {\bf context}, and below we will introduce the selected features under each category.
Each time series feature contains precisely $T$ values corresponding to the $T$ time steps.
For a concrete example of $\mathbf{X}$, see Figure~\ref{fig:teaser}.
All time series features are numerical, with the exception of the first one ({\small\texttt{\textbf{round\_type}}}), which is categorical. 
Each time step corresponds to a calendar month, and the steps are aligned monthly. 

\vspace{6pt}
\noindent\circled{\small\bf1}
{\bf Funding} category contains statistics of historical funding received by the company, showing recognition from investors.
\vspace{-3pt}
\begin{itemize}
    \item {\small\texttt{\textbf{round\_type}}} indicates the latest funding round type that a company has received up to time $t$, such as {\it Seed} or {\it Series A}, providing insights into its funding stage and maturity. 
    It is a categorical feature with 60 unique values.
    \item {\small\texttt{\textbf{total\_funding}}} is the cumulative amount of funding in USD that the company has received up to time $t$, indicating the amount of capital it has been able to attract. 
    The value range is from 1 to approximately $2\times 10^{11}$.
    \item {\small\texttt{\textbf{valuation}}}: the estimated USD valuation of the company immediately after its latest funding round and is included to provide insight into a company's overall financial value. 
    It is a numerical feature with values ranging from 1 to about $1\times 10^{12}$.
\end{itemize}
\vspace{-3pt}
\noindent\circled{\small\bf2}
{\bf Founder/Owner}: this category captures attributes of the founding team, which are critical to a company's short-term success and long-term survival \cite{c6_ghassemi2020automated}.
\vspace{-18pt}
\begin{itemize}
    \item {\small\texttt{\textbf{n\_founder}}} shows the number of a company's founders still with the company at time $t$.
    The value ranges from 0 to 38.
\end{itemize}
\vspace{-3pt}
\noindent\circled{\small\bf3}
{\bf Team}: this category captures the statistics of the employees of the company.
\vspace{-18pt}
\begin{itemize}
    \item {\small\texttt{\textbf{n\_employee}}}: the number of employees at time $t$, implying the company's growth trajectory. 
    The feature has a value range of 1 to 113,757.
\end{itemize}
\noindent\circled{\small\bf4}
{\bf Investor} category captures the statistics of investors who have funded the company, indicating its early attractiveness.
\vspace{-6pt}
\begin{itemize}
    \item {\small\texttt{\textbf{n\_investor}}} represents the total number of unique investors who have provided funding to the company up to time $t$. This feature provides insights into the diversification of the company's investment sources. 
    The value ranges from 1 to 240.
    \item {\small\texttt{\textbf{growth\_investor\_rate}}} is the ratio of unique GC investors\footnote{GC investors are defined as those who have participated in a funding round of 50 million USD or valuation above 200 million USD} among the company's unique investors up to time $t$. This feature indicates the investors' beliefs in the company's future growth potential.
    \item {\small\texttt{\textbf{average\_cagr}}} is the average Compound Annual Growth Rate (CAGR)\footnote{CAGR=(EV/SV)$^{1/\text{Y}}$-1 is calculated for each deal the investor has exited, where SV and EV stand for the starting and exiting value of the investment, respectively; Y is the number of holding years (from investment till divestment) of the invested asset.} of all exited deals made by the company's investors up to time $t$. This feature is meant to demonstrate the past investment performance of the involved investors.
    \item {\small\texttt{\textbf{2x\_cagr\_rate}}} is calculated as the ratio of investment deals up to time $t$ with a CAGR$\geq$2 among all exited deals made by the company's investors. This feature reflects the proportion of investors with a history of impressive returns who are currently invested in the company.
\end{itemize}
\noindent\circled{\small\bf5}
{\bf Web}: this category covers any feature extracted from web pages that are related to the company in focus.
\vspace{-6pt}
\begin{itemize}
    \item {\small\texttt{\textbf{cu\_popularity}}} describes the company's domain name popularity rank at time $t$ . This rank is determined based on the domain's network traffic as measured by Cisco Umbrella (CU)\footnote{\url{http://s3-us-west-1.amazonaws.com/umbrella-static/index.html}}.
    \item {\small\texttt{\textbf{sw\_global\_rank}}} describes at time $t$ the monthly unique visitors and pageviews of the company website(s). 
    The higher this sum, the higher the site's rank. 
    This feature is obtained from SimilarWeb\footnote{\url{https://support.similarweb.com/hc/en-us/articles/213452305-Rank}}.
    \item {\small\texttt{\textbf{n\_desktop\_visitor}}} and {\small\texttt{\textbf{n\_mobile\_visitor}}} are two features indicating the number of unique visitors to the company's website utilizing a desktop and mobile device, respectively. Both are sourced from SimilarWeb.
    \item {\small\texttt{\textbf{n\_news}}} counts the number of times a company is mentioned across approximately 3,700 news websites to a time point $t$, reflecting its media visibility and recognition.
    The value range of the dataset is 1 to 389.
\end{itemize}
\noindent\circled{\small\bf6}
{\bf Context}: this category captures extrinsic factors\footnote{While intrinsic features act from within a company, extrinsic ones wield their influence from the outside. The company may impact the former, yet not the latter.} that may be (but are not limited to) competition, regional, environmental, cultural or economical based.
\vspace{-2pt}
\begin{itemize}
    \item {\small\texttt{\textbf{n\_regional\_seed\_round}}} represents the number of seed funding rounds in the company's region\footnote{A region is a collection of countries such as \textit{Great Britain}, \textit{DACH}, \textit{France Benelux}, \textit{Southern Europe}, \textit{Nordics}, \textit{South Asia}, \textit{South East Asia}, and so on.} between adjacent time points $t\!-\!1$ and $t$. This feature offers context on the company's performance relative to regional competitors and financial conditions, highlighting potential company success even if regional investments are low.
    \item {\small\texttt{\textbf{n\_regional\_series\_ab}}}: same as the previous one except that it is counting the Series A and B rounds instead.
\end{itemize}


\subsection{TMTSC Architecture}
\label{sec:tmtsc}
\begin{wrapfigure}{r}{0.55\textwidth}
\vspace{-23pt}
  \begin{center}
    \includegraphics[width=\linewidth]{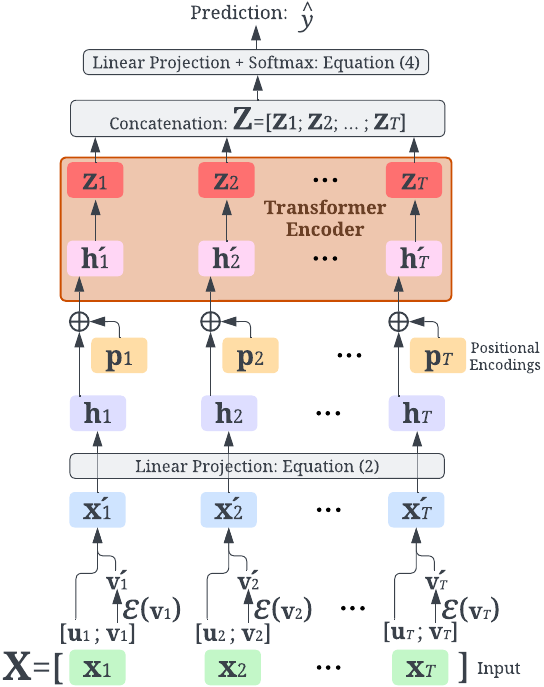}
  \end{center}
  \vspace{-12pt}
  \caption{TMTSC architecture: $\mathbf{u}_t$ and $\mathbf{v}_t$ are numerical and categorical part respectively.}
  \label{fig:architecture}
  \vspace{-19pt}
\end{wrapfigure}

As illustrated in Figure~\ref{fig:architecture}, TMTSC learns to predict $\hat{y}_n$ using time series input $\mathbf{X}$.
At the $t$-th time step, each input feature vector $\mathbf{x}_t$ consists of a numerical part (often normalized), denoted as $\mathbf{u}_t$, and a categorical part, denoted as $\mathbf{v}_t$. Thus, $\mathbf{x}_t=[\mathbf{u}_t; \mathbf{v}_t]$, where ``;'' represents a vector concatenation operation. 
To convert the categorical features $\mathbf{v}_t$ into dense embeddings, we utilize embedding layers, which can be collectively represented by a learnable function $\mathcal{E}$. 
The embedded categorical features are then given by $\mathbf{v}'_t = \mathcal{E}(\mathbf{v}_t)$.
As a result, the $K$-dimensional vector $\mathbf{x}_t$ is transformed to a new numerical vector $\mathbf{x}'_t$ that has $K'$ ($K'>K$) dimensions:
\begin{equation}
\label{eq:input}
\mathbf{x}'_t = [\mathbf{u}_t; \mathcal{E}(\mathbf{v}_t)]\in\mathbb{R}^{K'}\; \text{and }\; \mathbf{x}_t=[\mathbf{u}_t; \mathbf{v}_t]\in\mathbb{R}^K.
\end{equation}

Then, $\mathbf{x}'_t$ is linearly projected onto a $D$-dimensional vector space, where $D$ is the dimension of the Transformer model sequence element representations:
\begin{equation}
\label{eq:input-linear-transform}
\mathbf{h}_t = \mathbf{W}\mathbf{x}'_t+\mathbf{b},
\end{equation}
where $\mathbf{W}\in\mathbb{R}^{D\times K'}$ and $\mathbf{b}\in\mathbb{R}^D$ are learnable parameters and $\mathbf{h}_t\in\mathbb{R}^D$, $t\in\mathbb{Z}\cap[1,T]$ are the input vectors to the Transformer model.
Although Equation~\eqref{eq:input} and \eqref{eq:input-linear-transform} show the operation for a single time step for clarity, all raw input vectors $\mathbf{x}_t$, $t\in\mathbb{Z}\cap[1,T]$ are embedded in the same way concurrently.
It is worth mentioning that the above formulation can also accommodate univariate time series (i.e.,~$K=1$), though in the scope of this work, we will only evaluate the approach on multivariate time series.

It is important to note that the Transformer is a feed-forward architecture that does not inherently account for the order of input elements. 
To address the sequential nature of time series data, we incorporate positional encodings, denoted as $\mathbf{P}\!\!=\!\![\mathbf{p}_1, \mathbf{p}_2, \ldots, \mathbf{p}_T]\!\!\in\!\!\mathbb{R}^{T\times D}$, to the input vectors $\mathbf{H}\!=\![\mathbf{h}_1, \mathbf{h}_2, \ldots, \mathbf{h}_T]\!\in\!\mathbb{R}^{T\times D}$, resulting in the final input $\mathbf{H}'$:
\begin{equation}
\label{eq:final-input}
\mathbf{H}' = \mathbf{H}+\mathbf{P}=[\mathbf{h}'_1, \mathbf{h}'_2, \ldots, \mathbf{h}'_T]\in\mathbb{R}^{T\times D},
\end{equation}
where $\mathbf{h}'_t\in\mathbb{R}^D=\mathbf{h}_t+\mathbf{p}_t$. 
Closely following the approach in \cite{zerveas2021transformer}, we employ fully learnable positional encodings, as they have been reported to yield better performance compared to deterministic sinusoidal encodings \cite{vaswani2017attention} for multivariate time series classification tasks.
We also utilize batch normalization (rather than layer normalization), as it is considered effective in mitigating the impact of outlier values in time series data, an issue that does not arise for textual inputs.

The Transformer-based model architecture depicted in Figure~\ref{fig:architecture} generates $T$ output vectors $z_t$ corresponding to the $T$ input time steps. These output vectors are concatenated to form a single output matrix $\mathbf{Z}=[\mathbf{z}_1; \mathbf{z}_2; \ldots; \mathbf{z}_T]$, which serves as the input for a linear layer. As shown in Equation~\eqref{eq:final-output}, the linear layer is parameterized by $\mathbf{W}_{\text{out}}\in\mathbb{R}^{C\times(T\cdot D)}$ and $\mathbf{b}_{\text{out}}\in\mathbb{R}^C$, where $C$ denotes the number of classes to be predicted.
\begin{equation}
\label{eq:final-output}
\hat{\mathbf{y}} = \text{\it Softmax}(\mathbf{W}_{\text{out}}\mathbf{Z}+\mathbf{b}_{\text{out}}).
\end{equation}

\subsection{Optimization Target}
\label{sec:opt-target}
In the absence of a universally agreed-upon definition of ``true success'' of startups and scaleups, most existing definitions tend to focus on ``growth'', which can be measured from various perspectives, such as funding, revenue, employee count, and valuation, among others \cite{cao2022using}. As a well-established investment firm, we have access to a large volume of expert evaluations (akin to \cite{j26_bai2021startup,j37_kinne2021predicting}) that represent quantified assessments from human experts.
These evaluations encompass multiple categories/terms, such as ``inbound'', ``reviewing'', ``reach-out'', ``follow'', ``negotiating'', and ``out-of-scope''\footnote{The complete evaluation framework is withheld as it is proprietary.}, which are updated periodically by investment professionals for companies in the context of VC and GC. 
To further simplify the prediction task, we assign each evaluation term to either a good (``\verb|1|'') or bad (``\verb|0|'') binary bucket denoted by $\mathbf{y}_n$ in Figure~\ref{fig:teaser}, implying $C$=2 in Equation~\eqref{eq:loss}. 
In this way, each company is annotated with two ground-truth binary labels -- one for VC and the other for GC; and the loss function $\mathcal{L}$ is 
\begin{equation}
\label{eq:loss}
\mathcal{L} = - \frac{1}{N} \sum_{n=1}^N\left[ \mathbf{y}_n\log(\hat{\mathbf{y}}_n) + (1-\mathbf{y}_n)\log(1-\hat{\mathbf{y}}_n) \right].
\end{equation}

\begin{table}[t]
\addtolength{\tabcolsep}{4.5pt}
\renewcommand{\arraystretch}{1}
\footnotesize
\centering
\caption{Specification of tasks/datasets: split with an investor-centric strategy \cite{cao2022using}.}
\label{table:datasets_overview}
\begin{tabular}{l|rrrrrrr}               
\hline 
\textbf{Dataset} & \#feat. & \#sample & \#time step & \#class & \#train & \#val. & \#test \\                        
    \hline VC & 16 & 86,886 & 24 & 2 & 63,562 & 11,178 & 12,146 \\    
    \hline GC & 16 & 21,163 & 24 & 2 & 16,275 & - & 4,888\\  
\hline
\end{tabular}
\end{table}

\section{Experiments on Real-World Investment Tasks}
Following the details introduced in the previous section, we prepare two real-world proprietary datasets: ``{\bf VC}'' for the VC context and ``{\bf GC}'' for the GC context, performing data augmentation to obtain monthly time steps.
To eliminate overly sparse time series, we discard the samples whose time series features are all shorter than six months. Missing {\small\texttt{\textbf{valuation}}} values are approximated with the cumulative funding received up to that point. 
Missing {\small\texttt{\textbf{total\_funding}}} values are filled by taking the value of the previous month (if available) or 0 otherwise.
For the time steps where the values are still missing, we fill them with ``-1''.
Finally, we pad all time series to the same length of 24 months.
As for scaling, we empirically apply log-transform to 13 numerical features (excluding {\small\texttt{\textbf{cu\_popularity}}} and {\small\texttt{\textbf{n\_employee}}}).
The specification is presented in Table~\ref{table:datasets_overview}.
It is worth noting that we also experimented with two public TSC (time series classification) benchmark datasets\footnote{The overall performance can be found on Motherbrain's blog post: {\scriptsize\url{https://motherbrain.ai/applying-transformers-to-score-potentially-successful-startups-7893284efb01}}.}: Ethanol \cite{large2018detecting} and PEMS-SF \cite{cuturi2011fast}. Since they do not directly relate to the investment business domain, we chose to leave them outside the scope of this paper. For in-depth information about experiments on public datasets, we recommend reading~\cite{gustaf2023transformer}.

\begin{figure}[t!]
\centerline{\includegraphics[width=\linewidth]{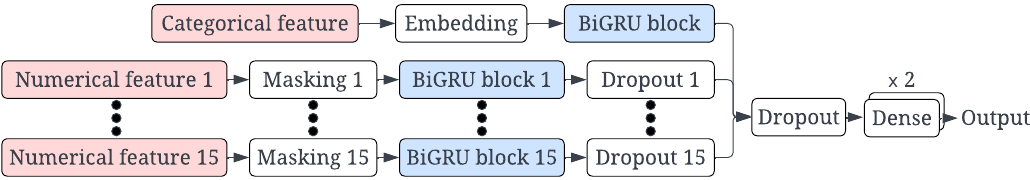}}
\vspace{-5pt}
\caption{\label{fig:ugru}U-GRU: each univariate time series is modeled by a BiGRU block.} 
\end{figure}

\begin{figure}[t!]
\centerline{\includegraphics[width=.85\linewidth]{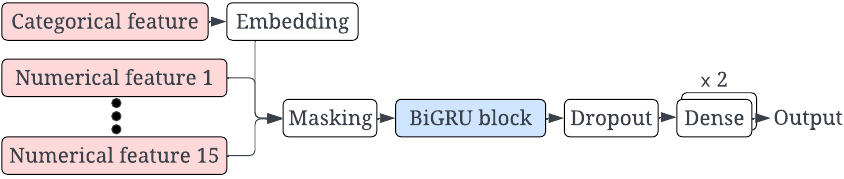}}
\vspace{-5pt}
\caption{\label{fig:mgru}M-GRU: all time series features are modeled by one single BiGRU block.} 
\end{figure}

\begin{figure}[t!]
\centerline{\includegraphics[width=0.85\linewidth]{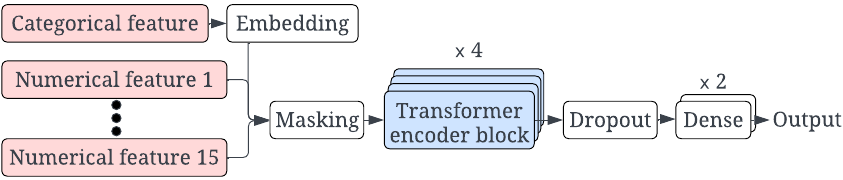}}
\vspace{-5pt}
\caption{\label{fig:transformer-encoder}TE architecture with 4 Transformer encoder blocks.} 
\vspace{-5pt}
\end{figure}

\subsection{Baselines and Hyper-Parameters}
GRU (Gated Recurrent Unit) \cite{chung2014empirical} is a highly relevant baseline for comparison due to its ability to model sequential dependencies and capture long-term dependencies in time series data.
We experiment both U-GRU (Univariate GRU) and M-GRU (Multivariate GRU), whose architectures are illustrated in Figure~\ref{fig:ugru} and \ref{fig:mgru} respectively.
We adopt the implementation of BiGRU (Bidirectional GRU) \cite{cho-etal-2014-learning}, masking, embedding, dropout, and dense layers from Keras\footnote{Keras Layer documentation: \url{https://keras.io/api/layers}}.
In both architectures, the masking layer is added to inform the model to ignore any values marked as missing in its computation. 
As a close relative and foundation of TMTSC, the Transformer Encoder (TE) \cite{vaswani2017attention} is also selected as a baseline.
As shown in Figure~\ref{fig:transformer-encoder}, we adopt the same layers as the original implementation \cite{vaswani2017attention}.
For comparability, the input features are ingested, embedded and concatenated in the same way as M-GRU as shown in Figure~\ref{fig:mgru}.
The hyper-parameters are selected based on the highest AUC-ROC (``Area Under the Curve'' of the Receiver Operating Characteristic curve) score on the validation split of the dataset. Refer to \cite{gustaf2023transformer} for the searched and selected hyper-parameter values.

\begin{table}[t!]
\begin{minipage}{.5\textwidth}
  \centering
\scriptsize
\renewcommand{\arraystretch}{1.1}
\addtolength{\tabcolsep}{-1pt}
\begin{tabular}{
>{\raggedright\arraybackslash}p{.65cm}|
p{1.35cm}| 
>{\raggedleft\arraybackslash}p{1cm}|
>{\raggedleft\arraybackslash}p{1cm}|
>{\raggedleft\arraybackslash}p{0.9cm}|
>{\raggedleft\arraybackslash}p{1cm}}
\hline
\textbf{Task} & Metric & U-GRU & M-GRU & TE & TMTSC \\
\hline
\parbox[t]{8mm}{\multirow{6}{*}{\textbf{VC}}}
& Accuracy $\pm$STDEV
& 0.548 {\bf $\pm$0.013}
& 0.731 $\pm$0.026 
& 0.655 $\pm$0.081 
& \textbf{0.863} $\pm$0.015  \\
\cline{2-6}
& Precision $\pm$STDEV 
& 0.704 {\bf $\pm$0.015} 
& 0.740 $\pm$0.044 
& 0.699 $\pm$0.062 
& \textbf{0.864} $\pm$0.016 \\
\cline{2-6}
& AUC-ROC $\pm$STDEV
& 0.628 $\pm$0.015 
& 0.819 $\pm$0.020 
& 0.780 $\pm$0.081 
& \textbf{0.924} {\bf $\pm$0.009} \\
\hline
\parbox[t]{8mm}{\multirow{6}{*}{\textbf{GC}}}
& Accuracy $\pm$STDEV
& 0.934 $\pm$0.011
& 0.924 $\pm$0.027 
& 0.933 $\pm$0.021 
& \textbf{0.956} {\bf $\pm$0.004}  \\
\cline{2-6}
& Precision $\pm$STDEV 
& 0.701 $\pm$0.058 
& 0.794 $\pm$0.101 
& 0.765 $\pm$0.108 
& \textbf{0.831} {\bf $\pm$0.026} \\
\cline{2-6}
& AUC-ROC $\pm$STDEV
& \textbf{0.977} $\pm$0.008 
& 0.939 $\pm$0.002 
& 0.971 $\pm$0.002 
& 0.971 {\bf $\pm$0.001} \\
\hline
\end{tabular}
\vspace{1pt}
\caption{\label{tab:main-results}
Overall performance comparison.}
\end{minipage}
\hspace{15pt}
\begin{minipage}{.44\textwidth}
  \centering
\scriptsize
\addtolength{\tabcolsep}{0.7pt}
\renewcommand{\arraystretch}{1.2}
\begin{tabular}{l|c|r|r}
\hline
\textbf{Task} & Method & Sec./Step & Relative Time \\
\hline
\parbox[t]{8mm}{\multirow{4}{*}{\textbf{VC}}} 
& U-GRU
& 2.000 
& 83.3 $\times$  \\
& M-GRU
& \underline{0.024}
& 1.0 $\times$ \\
& TE
& 0.057 
& 2.4 $\times$   \\
& TMTSC
& 0.100 
& 4.2 $\times$   \\
\hline
\parbox[t]{8mm}{\multirow{4}{*}{\textbf{GC}}} 
& U-GRU
& 1.232
& 46.6 $\times$  \\
& M-GRU
& \underline{0.026}
&  1.0 $\times$ \\
& TE
& 0.056 
& 2.1 $\times$   \\
& TMTSC
& 0.101
& 3.8 $\times$   \\
\hline
\end{tabular}
\vspace{2pt}
\caption{\label{tab:efficiency-results}
The comparison of training efficiency. Underlined values indicate shortest time per step for each dataset.}
\end{minipage}
\vspace{-2pt}
\end{table}

\subsection{Overall Performance: a Precision-Centric Comparison}
\begin{figure}[t!]
\centering
\subfigure[VC dataset\label{fig:roc-eqt-vc}]{\includegraphics[height=4.2cm]{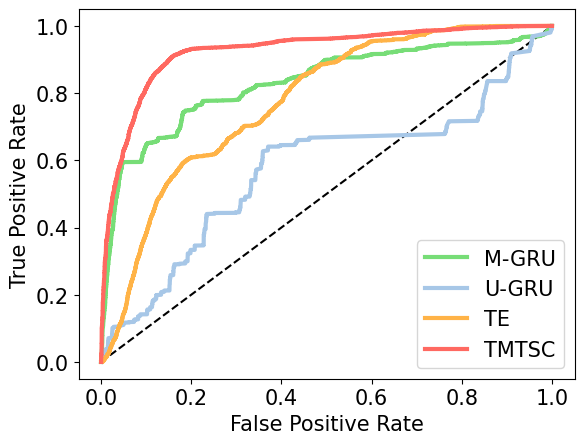}}
\hspace{12pt}
\subfigure[GC dataset\label{fig:roc-eqt-gc}]
{\includegraphics[height=4.2cm]{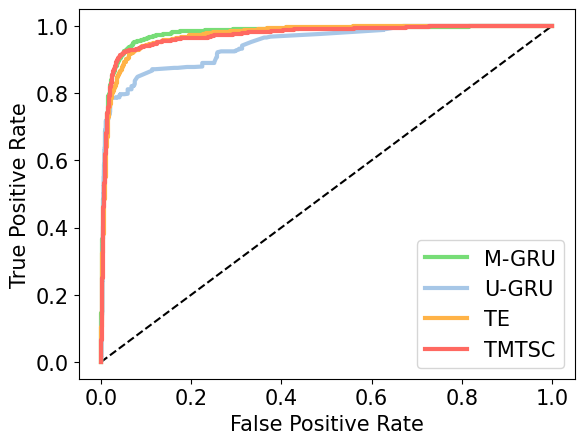}}
\vspace{-6pt}
\caption{ROC (Receiver Operating Characteristic) curves.} \label{fig:roc-curves}
\vspace{-6pt}
\end{figure}

When interpreting the results in Table~\ref{tab:main-results}, the costs of different types of prediction errors must be considered. The outcome from false negatives (failing to identify a successful company) is that investors are simply not made aware of a successful company and therefore no action is taken. In that regard, there is an upside loss in terms of lost profit but no detriment in terms of time or money invested. False positives (incorrectly predicting a company will be successful), on the other hand, can lead to wasted time spent on due diligence, or, in the worst case, an investment that loses money. For that reason, it is more important to evaluate a model with respect to its precision, or the number of its positive predictions that are actually positive.
Observing the precision scores in Table~\ref{tab:main-results}, TMTSC outperforms all other methods, achieving scores of 0.86 and 0.83 for VC and GC scenario, respectively.
Additionally, Figure~\ref{fig:roc-curves} provides a more balanced and comprehensive view using AUC-ROC metric. TMTSC clearly outperforms on the VC task, achieving an average score of 0.92, 12\% better than M-GRU, the next best method. 
All methods perform extremely well on GC task, and TMTSC's AUC-ROC score of 0.97 is less than 1\% lower than the winner U-GRU.

\subsection{Training Stability and Efficiency}
The standard deviation values (STDEV) in Table~\ref{tab:main-results} indicate that the TMTSC training is relatively stable in both the VC and GC datasets, as the values are relatively low compared to the other baselines.
To measure training efficiency, we record per-step training time for each dataset and method using the same batch size (=512) and hardware configurations. 
The results are presented in Table~\ref{tab:efficiency-results}, where the ``Relative Time'' column shows how the time consumption for the corresponding method relates to the fastest method (i.e.,~``1.0~$\times$''). Take the VC task for example, ``2.4~$\times$'' for TE would therefore mean TE took over twice as long as M-GRU.
It is evident that M-GRU requires the least amount of training time, largely due to its design, which favors simplicity. 
TMTSC and TE take only a small amount of extra time to train, despite their increased complexity. This is likely due to their multi-head architecture, allowing parallelization of self-attention computations.

\subsection{Portfolio Simulation}
\begin{figure}[t!]
\centering
\subfigure[VC dataset\label{fig:ps-eqt-vc}]{\includegraphics[width=6cm]{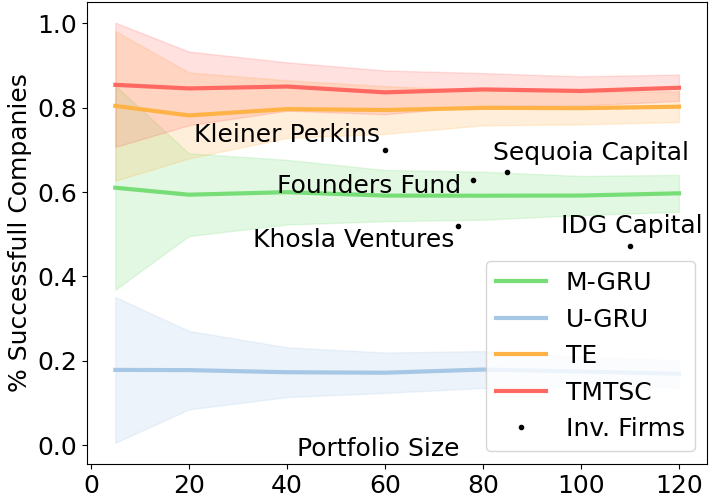}}
\subfigure[GC dataset\label{fig:ps-eqt-gc}]{\includegraphics[width=6cm]{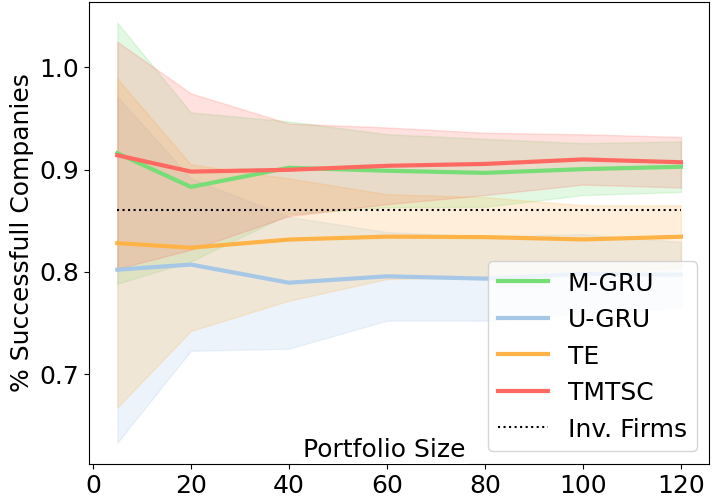}}
\vspace{-6pt}
\caption{Portfolio simulation: success rate vs. portfolio size.} \label{fig:portfolio-simulation}
\vspace{-5pt}
\end{figure}

To further evaluate the model in the context of the real-world investment scenario, portfolio simulations are executed and visualized in Figure~\ref{fig:portfolio-simulation}. 
Concretely, we assemble a set by isolating the companies confirmed to be potentially good investment targets in the VC or GC datasets (i.e.,~that are positively labeled).
From this set, we randomly sample $i$ companies to simulate forming VC/GC investment portfolios of size $i$ and calculate the percentage of companies each model predicts to be successful within the sample.
To address the stochasticity of this process, we perform each simulation 100 times.  Different portfolio sizes (i.e.,~values of $i$) are simulated; and for each $i$ (X-axis), the mean and standard deviations are plotted (Y-axis),
resulting in a colored line with a shaded area in Figure~\ref{fig:portfolio-simulation}.
In VC and GC contexts, we can see that (1) TMTSC performs the best among all methods, (2) performance becomes less variable as simulated portfolio size increases, and (3) the models evaluated performed more variably on the VC dataset than the GC dataset.

To roughly compare these methods against real-world VC and GC fund performance, Figure~\ref{fig:ps-eqt-vc}, includes the portfolio size and performance of five VC funds \cite{w2_yin2021solving} , showing a performance largely on par with M-GRU and inferior to TMTSC and TE.
For the GC simulation, a horizontal line representing the real-world GC success rate of 86.3\% \cite{mooradian2019growth} is included in Figure~\ref{fig:ps-eqt-vc} . Here, the real-world GC success rate is outperformed by M-GRU and TMTSC. 
It is important to note that investment firms are much more constrained than the simulation: they cannot invest in every attractive company they encounter due to factors like founders' preference, portfolio conflict, investment focus, and available funds.

\section{Conclusion and Future Work}
In this work, we propose using a Transformer-based Multivariate Time Series Classifier (TMTSC) to facilitate sourcing investment targets for Venture Capital (VC) and Growth Capital (GC). 
Specifically, TMTSC utilizes multivariate time series as input to predict the probability that any candidate company will succeed in the context of a VC or GC fund. 
We formally define the sourcing problem as a multivariate time series classification task,
and introduce the key components of our implementation, including
input features, model architecture, and optimization target.
Our extensive experiments on two proprietary datasets (collected from real-world VC and GC contexts) demonstrate the effectiveness, stability, and efficiency of our approach compared with three popular baselines.
To further evaluate the model in the context of the real-world investment scenario, portfolio simulations are executed, showing TMTSC's high success rate in both VC and GC sourcing.
The main future work includes (1) incorporating global features along with time series input, (2) and learning generic and condensed representations for multivariate time series for varies downstream prediction tasks.

\bibliographystyle{splncs04}
\bibliography{ref}

\begin{thebibliography}{10}
\providecommand{\url}[1]{\texttt{#1}}
\providecommand{\urlprefix}{URL }
\providecommand{\doi}[1]{https://doi.org/#1}

\bibitem{j16_allu2022predicting}
Allu, R., Padmanabhuni, V.N.R.: Predicting the success rate of a start-up using lstm with a swish activation function. Journal of Control and Decision  \textbf{9}(3),  355--363 (2022)

\bibitem{b1_ang2022using}
Ang, Y.Q., Chia, A., Saghafian, S.: Using machine learning to demystify startups’ funding, post-money valuation, and success. In: Innovative Technology at the Interface of Finance and Operations, pp. 271--296. Springer (2022)

\bibitem{arroyo2019assessment}
Arroyo, J., Corea, F., Jimenez-Diaz, G., Recio-Garcia, J.A.: Assessment of machine learning performance for decision support in venture capital investments. IEEE Access  \textbf{7},  124233--124243 (2019)

\bibitem{j26_bai2021startup}
Bai, S., Zhao, Y.: Startup investment decision support: Application of venture capital scorecards using machine learning approaches. Systems  \textbf{9}(3), ~55 (2021)

\bibitem{blank2013lean}
Blank, S.: Why the lean start-up changes everything. Harvard business review  \textbf{91}(5),  63--72 (2013)

\bibitem{block2019private}
Block, J., Fisch, C., Vismara, S., Andres, R.: {PE} investment criteria: An experimental conjoint analysis of venture capital, business angels, and family offices. Journal of Corporate Finance  \textbf{58},  329--352 (2019)

\bibitem{j17_bonaventura2020predicting}
Bonaventura, M., Ciotti, V., Panzarasa, P., Liverani, S., Lacasa, L., Latora, V.: Predicting success in the worldwide start-up network. Scientific Reports  \textbf{10}(1), ~1--6 (2020)

\bibitem{cao-etal-2023-ascalable}
Cao, L., von Ehrenheim, V., Berghult, A., Cecilia, H., Anselmo~Stahl, R., Wandborg, J., Stan, S., Catovic, A., Ferm, E., Hannes, I.: A scalable and adaptive system to infer the industry sectors of companies: Prompt + model tuning of generative language models. In: Proceedings of the IJCAI Workshop on Financial Technology And Natural Language Processing (FinNLP). pp. 1--11 (Aug 2023)

\bibitem{cao2022using}
Cao, L., von Ehrenheim, V., Krakowski, S., Li, X., Lutz, A.: Using deep learning to find the next unicorn: A practical synthesis. arXiv preprint arXiv:2210.14195  (2022)

\bibitem{cao-etal-2022-sire}
Cao, L., Horn, S., von Ehrenheim, V., Stahl, R.A., Landgren, H.: Simulation-informed revenue extrapolation with confidence estimate for scaleup companies using scarce time series data. In: Proceedings of the 31st ACM International Conference on Information and Knowledge Management (CIKM ’22), October 17–21, 2022, Atlanta, GA, USA. p. 12 pages. Association for Computing Machinery (ACM), New York, NY, USA (Oct 2022)

\bibitem{cavallo2019fostering}
Cavallo, A., Ghezzi, A., Dell'Era, C., Pellizzoni, E.: Fostering digital entrepreneurship from startup to scaleup: The role of venture capital funds and angel groups. Technological Forecasting and Social Change  \textbf{145},  24--35 (2019)

\bibitem{c16_chen2021trend}
Chen, M., Wang, C., Qin, C., Xu, T., Ma, J., Chen, E., Xiong, H.: A trend-aware investment target recommendation system with heterogeneous graph. In: Intl. Joint Conf. on Neural Networks. pp.~1--8 (2021)

\bibitem{c21_cheng2019success}
Cheng, C., Tan, F., Hou, X., Wei, Z.: Success prediction on crowdfunding with multimodal deep learning. In: IJCAI. pp. 2158--2164 (2019)

\bibitem{chernenko2021mutual}
Chernenko, S., Lerner, J., Zeng, Y.: Mutual funds as venture capitalists? evidence from unicorns. The Review of Financial Studies  \textbf{34}(5),  2362--2410 (2021)

\bibitem{cho-etal-2014-learning}
Cho, K., van Merri{\"e}nboer, B., Gulcehre, C., Bahdanau, D., Bougares, F., Schwenk, H., Bengio, Y.: Learning phrase representations using {RNN} encoder{--}decoder for statistical machine translation. In: Proceedings of the Conference on Empirical Methods in Natural Language Processing ({EMNLP}). pp. 1724--1734. Association for Computational Linguistics, Doha, Qatar (Oct 2014)

\bibitem{chung2014empirical}
Chung, J., Gulcehre, C., Cho, K., Bengio, Y.: Empirical evaluation of gated recurrent neural networks on sequence modeling. In: NIPS 2014 Workshop on Deep Learning, December 2014 (2014)

\bibitem{cuturi2011fast}
Cuturi, M.: Fast global alignment kernels. In: Proceedings of the 28th international conference on machine learning (ICML-11). pp. 929--936 (2011)

\bibitem{c9_dellermann2021finding}
Dellermann, D., Lipusch, N., Ebel, P., Popp, K.M., Leimeister, J.M.: Finding the unicorn: Predicting early stage startup success through a hybrid intelligence method. In: International Conference on Information Systems (2021)

\bibitem{w6_garkavenko2022you}
Garkavenko, M., Gaussier, E., Mirisaee, H., Lagnier, C., Guerraz, A.: Where do you want to invest? predicting startup funding from freely, publicly available web info. arXiv:2204.06479  (2022)

\bibitem{w3_gastaud2019varying}
Gastaud, C., Carniel, T., Dalle, J.M.: The varying importance of extrinsic factors in the success of startup fundraising: competition at early-stage and networks at growth-stage. arXiv:1906.03210  (2019)

\bibitem{c6_ghassemi2020automated}
Ghassemi, M., Song, C., Alhanai, T.: The automated venture capitalist: Data and methods to predict the fate of startup ventures. In: AAAI Workshop on Knowledge Discovery from Unstructured Data in Financial Services (2020)

\bibitem{gompers2020venture}
Gompers, P.A., Gornall, W., Kaplan, S.N., Strebulaev, I.A.: How do venture capitalists make decisions? Journal of Financial Economics  \textbf{135}(1),  169--190 (2020)

\bibitem{goodfellow2016deep}
Goodfellow, I., Bengio, Y., Courville, A.: Deep learning. MIT press (2016)

\bibitem{w5_guerzoni2019survival}
Guerzoni, M., Nava, C.R., Nuccio, M.: The survival of start-ups in time of crisis: a {ML} approach to measure innovation. arXiv preprint arXiv:1911.01073  (2019)

\bibitem{gustaf2023transformer}
Halvardsson, G.: A Transformer-Based Scoring Approach for Startup Success Prediction. Master's thesis, KTH Royal Institute of Technology (2023)

\bibitem{t2_horn2021deep}
Horn, S.: Deep learning models as decision support in venture capital investments: Temporal representations in employee growth forecasting of startup companies. Master's thesis, KTH Royal Institute of Technology \& EQT Partners (2021)

\bibitem{j30_kaiser2020value}
Kaiser, U., Kuhn, J.M.: The value of publicly available, textual and non-textual information for startup performance prediction. Journal of Business Venturing Insights  \textbf{14},  e00179 (2020)

\bibitem{j37_kinne2021predicting}
Kinne, J., Lenz, D.: Predicting innovative firms using web mining and deep learning. PloS One  \textbf{16}(4),  e0249071 (2021)

\bibitem{krishna2016predicting}
Krishna, A., Agrawal, A., Choudhary, A.: Predicting the outcome of startups: less failure, more success. In: International Conference on Data Mining Workshop. pp. 798--805 (2016)

\bibitem{large2018detecting}
Large, J., Kemsley, E.K., Wellner, N., Goodall, I., Bagnall, A.: Detecting forged alcohol non-invasively through vibrational spectroscopy and machine learning. In: Pacific-Asia Conference on Knowledge Discovery and Data Mining. pp. 298--309. Springer (2018)

\bibitem{w1_lyu2021graph}
Lyu, S., Ling, S., Guo, K., Zhang, H., Zhang, K., Hong, S., Ke, Q., Gu, J.: Graph neural network based vc investment success prediction. arXiv preprint arXiv:2105.11537  (2021)

\bibitem{j33_malmstrom2020they}
Malmstr{\"o}m, M., Voitkane, A., Johansson, J., Wincent, J.: What do they think and what do they say? gender bias, entrepreneurial attitude in writing and venture capitalists’ funding decisions. Journal of Business Venturing Insights  \textbf{13},  e00154 (2020)

\bibitem{mooradian2019growth}
Mooradian, P., Auerback, A., Slotsky, C., Gilfix, J.: Growth equity: Turns out, it’s all about the growth (2019)

\bibitem{raghu2017expressive}
Raghu, M., Poole, B., Kleinberg, J., Ganguli, S., Sohl-Dickstein, J.: On the expressive power of deep neural networks. In: International Conference on Machine Learning. pp. 2847--2854. PMLR (2017)

\bibitem{j8_santisteban2021critical}
Santisteban, J., Mauricio, D., Cachay, O., et~al.: Critical success factors for technology-based startups. International Journal of Entrepreneurship and Small Business  \textbf{42}(4),  397--421 (2021)

\bibitem{j43_shi2021leveraging}
Shi, J., Yang, K., Xu, W., Wang, M.: Leveraging deep learning with audio analytics to predict the success of crowdfunding projects. The Journal of Supercomputing  \textbf{77}(7),  7833--7853 (2021)

\bibitem{j22_skawinska2020success}
Skawi{\'n}ska, E., Zalewski, R.I.: Success factors of startups in the {EU} - a comparative study. Sustainability  \textbf{12}(19), ~8200 (2020)

\bibitem{t1_stahl2021leveraging}
Stahl, R.H.A.: Leveraging Time-Series Signals for Multi-Stage Startup Success Prediction. Master's thesis, ETH Zurich \& EQT Partners (2021)

\bibitem{j9_tang2022deep}
Tang, Z., Yang, Y., Li, W., Lian, D., Duan, L.: Deep cross-attention network for crowdfunding success prediction. IEEE Transactions on Multimedia  (2022)

\bibitem{vaswani2017attention}
Vaswani, A., Shazeer, N., Parmar, N., Uszkoreit, J., Jones, L., Gomez, A.N., Kaiser, {\L}., Polosukhin, I.: Attention is all you need. Advances in neural information processing systems  \textbf{30} (2017)

\bibitem{williamson2002research}
Williamson, K.: Research methods for students, academics and professionals: Information management and systems. Elsevier (2002)

\bibitem{w2_yin2021solving}
Yin, D., Li, J., Wu, G.: Solving the data sparsity problem in predicting the success of the startups with machine learning methods. arXiv preprint arXiv:2112.07985  (2021), \url{https://arxiv.org/abs/2112.07985}

\bibitem{zbikowski2021machine}
{\.Z}bikowski, K., Antosiuk, P.: A machine learning, bias-free approach for predicting business success using crunchbase data. Information Processing \& Management  \textbf{58}(4),  102555 (2021)

\bibitem{zerveas2021transformer}
Zerveas, G., Jayaraman, S., Patel, D., Bhamidipaty, A., Eickhoff, C.: A transformer-based framework for multivariate time series representation learning. In: Proceedings of the 27th ACM SIGKDD Conference on Knowledge Discovery \& Data Mining. pp. 2114--2124 (2021)

\bibitem{c1_zhang2021scalable}
Zhang, S., Zhong, H., Yuan, Z., Xiong, H.: Scalable heterogeneous graph neural networks for predicting high-potential early-stage startups. In: ACM SIGKDD Conference on Knowledge Discovery and Data Mining. pp. 2202--2211 (2021)

\end{thebibliography}

\appendix
\end{document}